\documentclass{esannV2}
\usepackage[dvips]{graphicx}
\usepackage[latin1]{inputenc}
\usepackage{amssymb,amsmath,array}

\begin{document}
\newcommand{\trans}[1]{#1^{\mathsf{T}}}
\newcommand{\argmin}{\mathop{\mathrm{arg\;min}}}
\newcommand{\argmax}{\mathop{\mathrm{arg\;max}}}
\newcommand{\sgn}{\mathop{\mathrm{sgn}}}
\newcommand{\la}{\leftarrow}
\newcommand{\bs}{\boldsymbol}
\title{A Unified View of TD Algorithms\\
{\small Introducing \emph{Full-gradient TD} and \emph{Equi-gradient descent TD}}}
\author{Manuel Loth
\thanks{The first author gratefully acknowledges the support from 
\emph{Region Nord - Pas-de-Calais} and \emph{INRIA} (PhD grant).}
\; and Philippe Preux
\vspace{.3cm}\\
INRIA-Futurs - SequeL\\
Universit\'e de Lille - LIFL \\
Villeneuve d'Ascq - France
}

\maketitle

\begin{abstract}
  This paper addresses the issue of policy evaluation in Markov
  Decision Processes, using linear function approximation. It provides
  a unified view of algorithms such as \emph{TD($\lambda$)},
  \emph{LSTD($\lambda$)}, \emph{iLSTD}, \emph{residual-gradient TD}.
  It is asserted that they all consist in minimizing a gradient
  function and differ by the form of this function and their means of
  minimizing it.  Two new schemes are introduced in that framework:
  \emph{Full-gradient TD} which uses a generalization of the principle
  introduced in \emph{iLSTD}, and \emph{EGD TD}, which reduces the
  gradient by successive \emph{equi-gradient descents}.  These three
  algorithms form a new intermediate family with the interesting property
  of making much better use of the samples than \emph{TD} while keeping 
  a gradient descent scheme, which is useful for complexity issues and
  optimistic policy iteration.  
\end{abstract}

\section{The policy evaluation problem}

A \emph{Markov Decision Process} (MDP) describes a dynamical system
and an agent.  The system is described by its state $s \in
\mathcal{S}$. When considering discrete time, the agent can apply at
each time step an action $u \in \mathcal{U}$ which drives the system
to a state $s'=u(s)$ at the next time step. $u$ is generally
non-deterministic.

To each transition is associated a reward $r \in \mathcal{R}\subset \mathbb{R}$. 
A policy $\pi$ is a function that associates to any state of the system an
action taken by the agent.

Given a discount factor $\gamma$, the value function $v^\pi$ of a policy $\pi$ 
associates to any state the expected discounted sum of rewards received when
applying $\pi$ from that state for an infinite time: 
$$v^\pi(s_0) = \mathbb{E}\left(\sum_{t=0}^{\infty} 
\gamma^t r(s_t \xrightarrow{\pi(s_t)} s_{t+1})\right)$$

% Evaluating a policy is useful for finding an optimal policy $\pi^*$,
% which has the highest value on all states. Given $\pi$ and $v^\pi$,
% one can define a policy $\pi'$ greedy on $v^\pi$, that always choose
% the action that maximizes the expected value of the next state.
% $v^{\pi'}$ is necessarily greater or equal to $v^{\pi}$ on all states,
% and repeating this policy improvement scheme always converges to an
% optimal policy, as long as the value functions are sufficiently well
% approximated.

This paper addresses the evaluation of a policy by approximating the value 
function as a linear combination of fixed features, and estimating the 
coefficients from sampled trajectories (sequences of visited states and 
received rewards when starting from a certain state).

All the information on $v$ contained in a trajectory 
$s_0 \xrightarrow{r_0} s_1 \xrightarrow{r_1} \ldots \xrightarrow{r_{n-1}} s_n$
lies in the following system of Bellman equations:
$$
\left\{\begin{array}{lcl}
    v(s_0) &=& r_0 + \gamma v(s_1)\\
%    v(s_1) &=& r_1 + \gamma v(s_2)\\
    \ldots\\
    v(s_{n-1}) &=& r_{n-1} + \gamma v(s_n)\\
\end{array}\right.
$$
The equalities are abusive when the actions are not deterministic, but
averaging these equations converges to valid equations as the number
of samples tends to infinity.

The policy evaluation problem consists in finding a function that
satisfies the most this system (which may include several
trajectories).  This can be achieved in several ways.
In the following, all major methods are described in a single and 
simple framework:\\
\begin{tabular}{p{0.95\linewidth}}
  $\bullet$ define a gradient function $\mu$ of the observed transitions 
  and parameters;\\
  $\bullet$ update its value whenever a new transition is observed;\\
  $\bullet$ whenever needed, modify the parameters in order to reduce $\mu$,
  and then update its value.
\end{tabular}

  Section~\ref{sec:gradient} discusses the two currently used gradient
  functions and their meaning.  Section~\ref{sec:td} presents the TD
  algorithms -- \emph{TD($\lambda$)}~\cite{sutton-barto} and
  \emph{residual-gradient TD}~\cite{baird-residual} -- in that
  framework.  Section~\ref{sec:lstd} shows that
  \emph{LSTD($\lambda$)}~\cite{boyan-lstd} and
  \emph{LSPE($\lambda$)}~\cite{nedic-lspe} and their
  \emph{Bellman-residual} versions share the same kind of derivation.
  Section~\ref{sec:full-gradient} discusses a third family of
  algorithms that use an intermediate update scheme (\emph{full
    gradient}). It includes
  \emph{iLSTD}~\cite{geramifard-ilstd,geramifard-ilstd-lambda} and
  two algorithms introduced in this paper: \emph{Full-TD} and
  \emph{Equi-gradient descent TD}.  Section~\ref{sec:exp} presents
  experimentations made on the \emph{Boyan chain} MDP, which
  illustrate some of the benefits and drawbacks of each method.
  Finally, the conclusion discusses the potential advantages of the
  full gradient scheme for optimistic policy iteration.

  Complete proofs of the equivalences of these
  formulations with the original ones and derivation of the
  equi-gradient descent algorithm are exposed in
  \cite{loth-unified-tr, loth-egd-tr}.

\section{Fixed-point gradient vs. Bellman-residual gradient}
\label{sec:gradient}

The \emph{TD(0)} algorithm estimates $v$ iteratively by using its
current estimate $\hat v$ to approximate the right hand side of these
equations:
\begin{eqnarray*}
v(s_t) = r_{t} + \gamma v(s_{t+1}) &\Rightarrow&
  v(s_t) \simeq r_{t} + \gamma \hat v(s_{t+1})\\
&\Rightarrow& v(s_t)-\hat v(s_t)\simeq r_{t}
- \hat v(s_t) + \gamma\hat v(s_{t+1})
\end{eqnarray*}
and consequently updating 
$\hat v(s_t) \leftarrow \hat v(s_t) + \alpha 
\left(r_{t} - \hat v(s_t) + \gamma\hat v(s_{t+1})\right)$

\emph{TD($\lambda$)} averages such approximations of $v(s_t)$ on all
``dynamic programming ranks''. It can be seen as expanding the system 
to all implicit equations:
$$
\left\{\begin{array}{l}
    v(s_0) = r_0 \text{+} \gamma v(s_1) = r_0 \text{+} \gamma (r_1 \text{+} \gamma v(s_2)) 
    = \ldots = r_0 \text{+} \gamma(r_1 \text{+} \gamma (r_2 \text{+} \ldots \text{+} \gamma v(s_n)))\\
    v(s_1) = r_1 \text{+} \gamma v(s_2) = \ldots\\
    \ldots
\end{array}\right.
$$
and again replacing $v$ by $\hat v$ in the right hand sides. The different
estimations of $v(s_t)$ are averaged using coefficients determined by
a value $\lambda \in [0,1]$, which leads to estimating $v(s_t)-\hat
v(s_t)$ by $\sum_{\tau=t}^{T-1} (\lambda\gamma)^{\tau-t} (r_{\tau} -
\hat v(s_\tau) + \gamma \hat v(s_{\tau+1}))$.  This error signal is
again used to update $\hat{v}(s_t)$. In the case of linear
approximators, the vector of error signals on $\hat{v}(s_0), \ldots, \hat{v}(s_{T-1})$
can be written as $\mathbf{L}(\mathbf{r}-\mathbf{B}\bs{\Phi\omega})=$
$$
\left(\begin{array}{@{\;}c@{\;}c@{\;}c@{\;}c@{\;}}
    1&\lambda\gamma&(\lambda\gamma)^2&\ldots\\
    &1&\lambda\gamma&\ldots\\
    \mathbf{0}&&\ddots
\end{array}\right)
\left[
  \left(\begin{array}{@{\;}c@{\;}}
      r_0\\\vdots\\r_{T-1}
    \end{array}\right)
  -
  \left(\begin{array}{@{\;}c@{\;}c@{\;}c@{\;}c@{\;}}
    1&-\gamma&&\mathbf{0}\\
    &1&-\gamma\\
    \mathbf{0}&&\ddots
  \end{array}\right)
  \left(\begin{array}{c@{\;}c@{\;}c@{\;}c}
    \phi_1(s_0)&\ldots&\phi_n(s_0)\\
    \vdots&&\vdots\\
    \phi_1(s_T)&\ldots&\phi_n(s_T)\\
  \end{array}\right)
  \left(\begin{array}{@{\;}c@{\;}}
      \omega_1\\\vdots\\\omega_n
    \end{array}\right)
\right]
$$
They are projected on the parameter $\bs\omega$ of $\hat{v}$ by
$\trans{\bs{\Phi}}\mathbf{L}(\mathbf{r}-\mathbf{B}\bs{\Phi\omega})$.
This gives what one can call a fixed-point gradient, which is the sum of
these on all trajectories (\textit{ie.} the same expression with adequately 
extended vectors and matrices).

Another way of doing is to aim at solving the Bellman system,
\textit{ie.} minimize $\|\mathbf{r}-\mathbf{B}\bs{\Phi\omega}\|_2^2$
w.r.t. $\bs\omega$. This gives the Bellman-residual gradient
\mbox{$\trans{\bs{\Phi}}\trans{\mathbf{B}}(\mathbf{r}-\mathbf{B}\bs{\Phi\omega})$}.

The conceptual difference is simple: The fixed-point gradient
transforms the errors on transitions (temporal differences) on the
approximate value function itself (\textit{ie.} errors on single
states) by a multi-rank dynamic programming scheme, and then projects
these estimated errors on the parameter $\bs\omega$, whereas the
Bellman-residual gradient does a direct projection.

The iterative computation of these gradients proceeds according to the
following way: the components of the vector
$\mathbf{r}-\mathbf{B}\bs{\Phi\omega}$ are the successive temporal
differences $d_t=r_t - \hat{v}(s_t) + \gamma \hat{v}(s_{t+1})$; the
columns of $\trans{\bs\Phi}\mathbf{L}$ or
$\trans{\bs\Phi}\trans{\mathbf{B}}$ are referred to as the
\emph{eligibility traces} $\mathbf{z_t}$ in the first case -- this
denomination will be extended here to the second case.  Each new
sampled transition modifies the gradient $\bs\mu$ by
$\bs{\mu_t}\la\bs{\mu_{t-1}}+ d_t\mathbf{z_t}$, $\mathbf{z_t}$ itself
being computed iteratively.

These gradients, as well as $\hat{v}$, are linear in $\bs\omega$:
$\bs\mu = \mathbf{A}\bs\omega + \mathbf{b}$, with
$\mathbf{b}=\trans{\Phi}\mathbf{Lr}$, and
$\mathbf{A}=\trans{\bs\Phi}\mathbf{L}\mathbf{B}\bs\Phi$ in the
fixed-point case, or
$\mathbf{A}=\trans{\bs\Phi}\trans{\mathbf{B}}\mathbf{B}\bs\Phi$ in the 
Bellman-residual case.

In the following, let us note $\bs{\delta_\omega}$ the additive term of
any update of $\bs\omega$ in the algorithms.

\section{TD algorithms}
\label{sec:td}
\emph{TD($\lambda$)}~\cite{sutton-barto}, in its purely iterative
form, performs the following update after each transition:
$\bs\omega\la\bs\omega + \alpha d_t\mathbf{z_t}$. Equivalently, the
updates can be performed only after each trajectory, which is more
consistent with its definition. Depending on one's view (related to
the backward/forward views discussed in \cite{sutton-barto}), the
first scheme can be considered as the natural one and the second as
cumulating successive updates before commiting it at the end, or the
second one can be seen as more natural (given the explanation in the
previous section) and the first one as a partial update given the
partial computation of $\bs\mu$.  Note that here, $\bs\mu$ only
concerns the current trajectory: the updates performed in
\emph{TD($\lambda$)} only take into account the last trajectory.

Let us take a neutral point of view and state that the algorithm considers
the gradient on the current trajectory and update weights at any chosen time 
(but necesseraly including the end of the trajectory) by
$\bs\omega\la\bs\omega + \alpha \bs\mu$ followed by $\mu \la \mathbf{0}$: $\bs\mu$
is computed iteratively, and each time a partial computation has been used, it is
``thrown away''. At the end of each trajectory, the associated gradient has been used
for one update $\bs\omega\la\bs\omega + \alpha \bs\mu$ and is then forgotten.

To summarize, given the fixed-point gradient function $\mu(\textit{observed
transitions}, \bs\omega)$, \emph{TD($\lambda$)} updates $\bs\mu$ after
each transition (as exposed in previous section), and --whenever
wanted-- performs a parameter update $\bs\omega \la \bs\omega + \alpha
\bs\mu$ followed by $\bs\mu \la \mathbf{0}$.

The \emph{residual-gradient TD} algorithm~\cite{baird-residual} is
actually the same algorithm, only using the Bellman-residual
gradient.

\section{LSTD algorithms}
\label{sec:lstd}
It has been shown in \cite{tsitsiklis-analysis} that $\bs\omega$ converges in
\emph{TD($\lambda$)} to $\bs\omega^*$ such that
$\bs\mu(\bs\omega^*)=\mathbf{A}\bs\omega^*+\mathbf{b}=\mathbf{0}$. This
lead to the \emph{LSTD($\lambda$)} algorithm~\cite{boyan-lstd}
which, given sampled trajectories, directly computes 
$\bs\omega^* = \mathbf{A}^{-1}\mathbf{b}$.

For various motivations like numerical stability, use of optimistic policy
iteration, the possible singularity of $\mathbf{A}$, smooth processing
time, or getting a specific point of view on the algorithm, the
computation can be performed iteratively.
The algorithm can then be described as follows:\\
\begin{tabular}{p{0.95\linewidth}}
$\bullet$ for each new
transition, update $\bs\mu$ as exposed in section \ref{sec:gradient}, 
and update $\mathbf{A}^{-1}$ (using Shermann-Morrisson formula),\\
$\bullet$ whenever wanted, reduce $\bs\mu$ by updating $\bs\omega 
\la \bs\omega + \mathbf{A}^{-1}\bs\mu$. $\bs\omega$ is then the exact
solution of $\bs\mu(\textit{samples so far},\bs\omega)=\mathbf{0}$ and
$\bs\mu$ is updated to $\mathbf{0}$.\\
Again, the same algorithm can be applied using the Bellman-residual 
gradient.\\
\end{tabular}

\cite{nedic-lspe} introduced a similar algorithm, namely
\emph{Least Squares Policy Evaluation}. The difference resides in
updating $\bs\omega \la \bs\omega +
\left(\trans{\bs\Phi}\bs\Phi\right)^{-1}\bs\mu$,
and consequently updating $\bs\mu\la\bs\mu - \mathbf{A}
\bs{\delta_\omega}$.

\section{Full-gradient algorithms}
\label{sec:full-gradient}
Three algorithms are presented in this section that all rely on the same idea:
reduce $\bs\mu$ (again at any time) in a gradient descent way, but maintain
its ``real'' value: instead of zeroing it after each update, which corresponds
to forgetting each trajectory after only one gradient descent step on its
contribution to the overall gradient $\bs\mu$, the residual of the gradient
is kept, and thus the following updates not only perform one gradient descent
step on the current trajectory, but also continue this process for the previous
ones.

The first natural algorithm is introduced here as \emph{Full-gradient TD} and
consists in replacing $\bs\mu\la\mathbf{0}$ by 
$\bs\mu\la\bs\mu - \mathbf{A}\bs{\delta\omega}$ in the \emph{TD} algorithm.

The \emph{iLSTD} algorithm was introduced in
\cite{geramifard-ilstd,geramifard-ilstd-lambda} (as well as the
notation $\bs\mu$). Although it is presented as a variation of
\emph{LSTD} (hence its name), it is most related to gradient descent
than to the exact least-squares solving scheme. With the ``any-time
update'' generalization used throughout this article, it can be
described as a full-gradient TD in which $\bs\omega$ is updated only
on its more correlated component: $\omega_i\la\omega_i+\alpha\mu_i,$
\quad with $i=\argmax |\mu_i|$.

Finally, the equi-gradient descent (EGD) TD, introduced here, consists
in taking EGD~\cite{loth-egd-tr} steps as an update scheme. In a few
words, EGD also consists in modifying only the most correlated
parameter $\omega_i$, but $\alpha$ is chosen such that after this
update, another parameter $\omega_j$ becomes equi-correlated. The next update
is $\left(\begin{array}{c}\omega_i\\\omega_j\end{array}\right) \la
\left(\begin{array}{c}\omega_i\\\omega_j\end{array}\right)
+\alpha_2\left(\begin{array}{cc}A_{ii}&A_{ij}\\A_{ji}&A_{jj}\end{array}\right)^{-1}
\left(\begin{array}{c}\mu_i\\\mu_j\end{array}\right)$, and so on.
The constraint is that to allow the exact computations of the step lengths,
$\bs\mu$ must not be modified (by new samples) in between those steps. So a
typical update schedule is to perform a certain number of steps at the end
of each trajectory, preferably to one or a few steps after each transition.

The benefit exposed in the first paragraph comes at the cost of
maintaining the matrix $\mathbf{A}$, which has the same order of complexity as
maintaining $\mathbf{A}^{-1}$ in \emph{LSTD}, but is still about half less
complex.  However, as exposed in \cite{geramifard-ilstd}, if the features
are sparse (states have a non-zero value only on a subset of the
features), the complexity of the two last algorithms can be lowered,
unlike in \emph{LSTD}.

\emph{EGD TD} presents the crucial benefit of not having to tune the $\alpha$
update parameter of gradient descent schemes. Instead of setting the lengths
of descent steps \textit{a priori} and uniformely, and cross-validate them, 
they are computed on the fly given the data. 

\section{Experiments}
\label{sec:exp}
Experiments were run on a 100 states Boyan chain MDP~\cite{boyan-lstd}.
Details are exposed in \cite{loth-unified-tr}. 
The fixed-point gradient was used, with $\lambda=0.5$. Here are plotted\\
\begin{tabular}{p{0.95\linewidth}}
  $\bullet$ in \ref{Fig:traj}, the RMSE against the  number of 
  trajectories, which illustrates the differences between full exploitation
  of the samples (least-squares and full-gradient methods) and \emph{TD},\\
  $\bullet$ in \ref{Fig:time}, the RMSE against the computational time, 
  where the three families are clearly clustered. Note that the sparsity 
  of the features has not been taken into account, and \emph{EGD TD} and 
  \emph{iLSTD} can perform much better on that point, as experimented in 
  \cite{geramifard-ilstd} for the latter. 
\end{tabular}

\begin{figure}[h!]
\centering
\includegraphics[width=10cm, height=4.4cm]{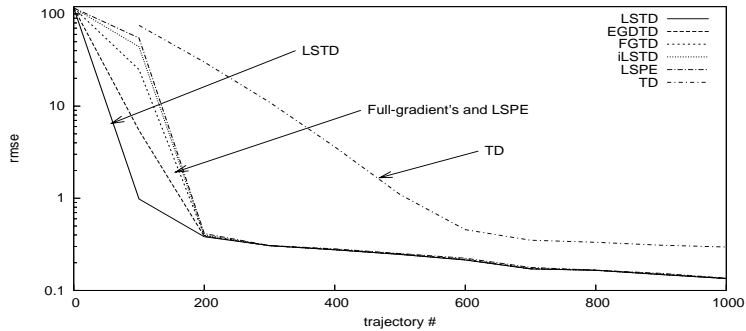}
\caption{Root mean squared error against the number of trajectories}
\label{Fig:traj}
\end{figure}
\begin{figure}[h!]
\centering
\includegraphics[width=10cm, height=4.4cm]{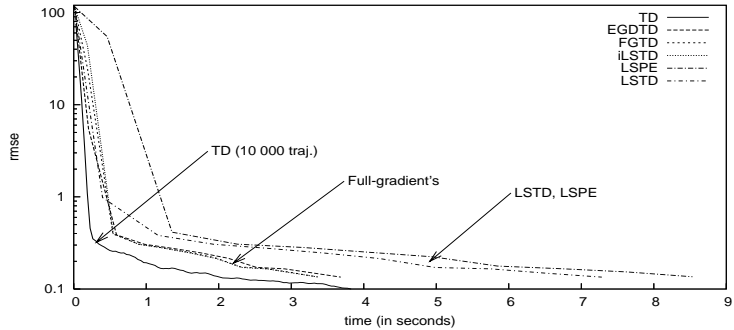}
\caption{Root mean squared error against the computational time}
\label{Fig:time}
\end{figure}
\section{Summary and perspectives}
The classical algorithms of reinforcement learning have been presented here
in a view both practical and enlightning. This view allows a natural
introduction of a new intermediate family of algorithms that performs
stochastic reduction of the errors, as in TD, but make full use of the
samples, as in LSTD. Let alone the time or sample complexity, these
methods open interesting perspectives in the frame of optimistic
policy iteration. Indeed, the principle of neither forgetting samples
after a small update, nor directly fully take them into account, may
allow to make a better use of samples than TD while avoiding the issue 
met by LSTD in that frame: making too much case of samples coming from
previous policies. This can be achieved by scaling $\bs\mu$ by a 
discount factor after each trajectory (for example), which amounts to
reducing only a given ratio of it.

\begin{footnotesize}
\bibliographystyle{unsrt}
\bibliography{rl}
\end{footnotesize}

\end{document}